\documentclass[runningheads]{llncs}
\usepackage{times}
\usepackage{url}
\usepackage{latexsym}
\usepackage{booktabs}
\usepackage{pbox}
\usepackage{amsmath}
\usepackage{amsfonts}
\usepackage{adjustbox}
\usepackage{array}
\usepackage{caption}

\usepackage[colorlinks,
            linkcolor=black,
            anchorcolor=blue,
            citecolor=blue,
            urlcolor=blue,
            ]{hyperref}

\usepackage[misc]{ifsym}
\usepackage[final]{changes}
\usepackage{caption, subcaption}

\begin{document}
\setlength{\abovedisplayskip}{4pt}      
\setlength{\belowdisplayskip}{5pt}
 \title{Multi-Perspective Fusion Network for Commonsense Reading Comprehension}
%
%
\author{Chunhua Liu\inst{1} \and Yan Zhao\inst{1} \and
Qingyi Si\inst{1} \and Haiou Zhang\inst{1} \and Bohan Li\inst{1} \and Dong Yu\inst{1,2} \Letter}
\authorrunning{C. Liu et al.}
%
\institute{Beijing Language and Culture University \and 
Beijing Advanced Innovation for Language Resources of BLCU
\email{\{chunhualiu596,zhaoyan.nlp\}@gmail.com\\
\{xk17sqy,hozhangel,yudong\_blcu\}@126.com\\
bohanli.lavida@gmail.com}
}
\maketitle              
\begin{abstract}
\deleted{
Commonsense Reading Comprehension (CRC) is a significantly challenging task in  reading comprehension, with the goal of making a right choice from two answer candidates for the questions referring to a narrative passage that may require commonsense knowledge. 
}

Commonsense Reading Comprehension (CRC) is a significantly challenging task, aiming at choosing the right answer for the question referring to a narrative passage, which may require commonsense knowledge inference. \deleted{the given narrative passage.} 
Most of the existing approaches only fuse the interaction information of choice, passage, and question in a simple combination manner from a \emph{union} perspective, which lacks the comparison information on a deeper level. 
Instead, we propose a Multi-Perspective Fusion Network (MPFN), extending the single fusion method with multiple perspectives by introducing the \emph{difference} and \emph{similarity} fusion\deleted{along with the \emph{union}}. More comprehensive and accurate information can be captured through the three types of fusion. We design several groups of experiments on MCScript dataset \cite{Ostermann:LREC18:MCScript} to evaluate the effectiveness of the three types of fusion respectively. From the experimental results, we can conclude that the difference fusion is comparable with union fusion, and the similarity fusion needs to be activated by the union fusion. The experimental result also shows that our MPFN model achieves the state-of-the-art with an accuracy of 83.52\% on the official test set. 

\end{abstract}
\section{Introduction}

\deleted{
\subsection{paragraph 1}
\emph{Content: Task Definition}
\begin{itemize}
\item [ ] 1. Describe the task of commonsense reading comprehension(CRC) belongs to which filed and how important it is.   
\item [ ] 2. Define the task of CRC
\item [ ] 3. Data feature of CRC 
\item [ ] 4. Figure 1 shows an example. 
\end{itemize}
}
Machine Reading Comprehension (MRC) is an extremely challenging topic in natural language processing field. 
It requires a system to answer the question referring to a given passage.\deleted{no matter whether the answer is mentioned in the passage.} 
\deleted{MRC consists of several sub-tasks, such as cloze-style reading comprehension, span-extraction reading comprehension, and open-domain reading comprehension. 
Most of existing datasets emphasize the question whose answer is mentioned in the passage since it does not need any commonsense. }
In real reading comprehension, the human reader can fully understand the passage with the prior knowledge to answer the question. 
To directly relate commonsense knowledge to reading comprehension, SemEval2018 Task ${11}$ defines a new sub-task called Commonsense Reading Comprehension, aiming at answering the questions that requires both commonsense knowledge and the understanding of the passage. The challenge of this task is how to\deleted{lies in} answer questions \deleted{requires a system to draw inferences from multiple sentences from the passage and requires}with the commonsense knowledge that does not appear in the passage explicitly. 
Table~\ref{table:example_MCS} shows an example of CRC. 
\begin{table}
\begin{tabular}{p{\columnwidth}}
\toprule
\textbf{Passage:} It was night time and it was time to go to bed. The boy wanted to keep playing. I told him that after he got ready for bed I would read a story to him. He dawdled a bit but finally started getting ready for bed.  First of all he had to take a bath. He splashed in the tub and split water all over the floor. Next he dried off in a big, fluffy blue towel.  Then he brushed his teeth with his special Star Wars toothbrush. Next he dressed in his Star Wars underwear and then put on his Star Wars pajamas.  His dad and I tucked him into his bed that was made with Star Wars sheets.  He said his prayers. Next was story time. I pulled out his favorite book about (you guessed it) Star Wars. He gradually dozed off dreaming about Anakin Skywalker and a galaxy far, far away. \\    
\vspace{0.1mm}

\textbf{Q1:} Did they sleep in the same room as their parents?\deleted{type="commonsense">} \\
A. Yes, they all slept in one big loft $\quad$ B. No they have their own room \\
\vspace{0.1mm} 
\textbf{Q2:} Why didn't the child go to bed by themselves? \deleted{" type="text">} \\
A. The child wanted to watch a Star Wars movie. $\quad$ B. The child wanted to continue playing.\\
\bottomrule
\end{tabular} \\[1em]  
\caption{\label{table:example_MCS} An example of CRC.}
\vspace{-20pt}
\end{table}



\deleted{
\subsection{paragraph 2}
\emph{Content: Previous Research}
\begin{itemize}
\item [ ] 1. Category the methods in SemEval2018 task 11
\item [ ] 2. Describe the first method
\item [ ] 3. Describe the second method 
\item [ ] 4. State that your work is belong to which method
\end{itemize}
}
Most studies on CRC task are neural network based (NN-based) models, which typically have the following characteristics. Firstly, word representations are augmented by additional lexical information. \deleted{, such as pre-trained embedding, POS and NER embedding, Relation embedding and some other handcraft features. }
Secondly, the interaction process is usually implemented by the attention mechanism, which can provide the interaction representations like choice-aware passage, choice-aware question, and question-aware passage. 
Thirdly, the original representations and interaction representations are fused together and then aggregated by a Bidirectional Long Short-Term Memory Network (BiLSTM)  \cite{Hochreiter:1997:LSTM} to get high-order semantic information. Fourthly, the final output based on their bilinear interactions. 
\deleted{
is the sum scores of passage to choice and question to choice. 
}

The NN-based models have shown powerfulness on this task. However, there are still some limitations.  
Firstly, the two fusion processes of passage and question to choice are implemented separately, until producing the final output. 
Secondly, the existing fusion method used in reading comprehension task is usually implemented by concatenation \cite{xiong:2018:dcn-plus,Chen:2018:HMA}, which is monotonous and cannot capture the partial comparison information between two parts. Studies on Natural Language Inference (NLI) have explored more functions \cite{Mou-EtAl:2016:P16-2,Chen-zhu:2017:Long3}, such as element-wise subtraction and element-wise multiplication, to capture more comparison information, which have been proved to be effective.  

In this paper, we introduce a Muti-Perspective Fusion Network (MPFN) to tackle these limitations. The model can fuse the choice with passage and question simultaneously to get a multi-perspective fusion representation.
Furthermore, inspired by the element-wise subtraction and element-wise multiplication function used in \cite{Chen-zhu:2017:Long3}, we define three kinds of fusion functions from multiple perspectives to fuse choice, choice-aware passage, and choice-aware question.
The three fusions are union fusion, difference fusion, and similarity fusion. Note that, we name the concatenation fusion method as union fusion in this paper, which collects the global information. The difference fusion and the similarity fusion can discover the different parts and similar parts among choice, choice-aware passage, and choice-aware question respectively.  
 

MPFN comprises an encoding layer,  a context fusion layer, and an output layer. In the encoding layer, we employ a BiLSTM as the encoder to obtain context representations. \deleted{to convert the embeddings of passage, question, and choice to their corresponding context embeddings.} To acquire better semantic representations, we apply union fusion in the word level. \deleted{to choice, choice-aware passage embedding, and choice-aware question embedding.} 
In the context fusion layer, we apply union fusion, difference fusion, and similarity fusion to obtain a multi-perspective fusion representation. In the output layer, a self-attention and a feed-forward neural network are used to make the final prediction.

We conduct experiments on MRScript dataset released by \cite{Ostermann:LREC18:MCScript}. Our single and ensemble model achieve
the accuracy of 83.52\% and 84.84\% on the official test set respectively. Our main contributions are as follows:  
\begin{itemize}
\item We propose a general fusion framework with two-layer fusion, 
which can fuse the passage, question, and choice simultaneously. 
\item To collect multi-perspective fusion representations, we define three types of fusions, consisting of union fusion, difference fusion, and similarity fusion.  
\deleted{
\item  We extend the fusion method to multi-perspective to obtain deeper understanding of the passage, question, and choice. 
}
\item We design several groups of experiments to evaluate the effectiveness of the three types of fusion and prove that our MPFN model outperforms all the other models. \deleted{ with an accuracy of 83.52\%.}
\end{itemize}

\section{Related Work}
MRC has gained significant popularity over the past few years. Several datasets have been constructed for testing the comprehension ability of a system, such as \emph{MCTest} \cite{Richardson:2013:MCTestAC}, \emph{SQuAD} \cite{Rajpurkar:2016:SQUAD}, \emph{BAbI} \cite{Weston:2015:BAbI}, \emph{TriviaQA} \cite{Joshi:2017:TriviaQA}, RACE \cite{Lai:2017:RACE}, and NewsQA \cite{Trischler:2017:NewsQA}. \deleted{The types of passage, question and answer of these datasets are various.} Each dataset focuses on one specific aspect of reading comprehension. Particularly, the MCScript \cite{Ostermann:LREC18:MCScript} dataset concerns answering the question which requires using commonsense knowledge.

\deleted{including Wikipedia articles, examinations, narrative stories, news articles.
Answering questions in these datasets.  
Meanwhile, the question types and answer types vary differently. The answer type 
multiple choice, span-answer, exact match}



Many architectures on MRC follow the process of representation, attention, fusion, and aggregation \cite{Seo:2016:BiDAF,xiong:2018:dcn-plus,Zhu:2018:RACE,Huang:2017:FusionNet,Wang:2017:r-net,Xu:2017:DFN}. BiDAF \cite{Seo:2016:BiDAF} fuses the passage-aware question, the question-aware passage, and the original passage in context layer by concatenation, and then uses a BiLSTM for aggregation. The fusion levels in current advanced models are categorized into three types by \cite{Huang:2017:FusionNet} , including word-level fusion, high-level fusion, and self-boosted fusion. They further propose a FusionNet to fuse the attention information from bottom to top to obtain a fully-aware representation for answer span prediction. 

\deleted{\cite{Xu:2017:DFN} present a DFN model to fuse the passage, question, and choice by dynamically determine the attention strategy. }

On SemEval2018 Task 11, most of the models use the attention mechanism to build interactions among the passage, the question, and the choice \cite{Wang:2018:TriAN,Chen:2018:HMA,Xia:2018:Jiangnan,Merkhofer:2018:MITRE}. The most competitive models are \cite{Wang:2018:TriAN,Chen:2018:HMA}, and both of them employ concatenation fusion to integrate the information. \cite{Wang:2018:TriAN}
utilizes choice-aware passage and choice-aware question to fuse the choice in word level. In addition, they apply the question-aware passage to fuse the passage in context level. Different from \cite{Wang:2018:TriAN}, both the choice-aware passage and choice-aware question are fused into choice in the context level in \cite{Chen:2018:HMA} , which is the current state-of-the-art result on the MCSript dataset.

On NLI task, fusing the premise-aware hypothesis into the hypothesis is an effective and commonly-used method. \cite{wang-jiang:2016:mlstm,Parikh:2016:DecomAtt} leverage the concatenation of the hypothesis and the hypothesis-aware premise to help improve the performance of their model.  The element-wise subtraction and element-wise multiplication between the hypothesis and the hypothesis-aware premise are employed in \cite{Chen-zhu:2017:Long3} to enhance the concatenation. \deleted{and further achieved the state-of-the-art results on Stanford Natural Language Inference \cite{snli:emnlp2015} benchmark.} 

Almost all the models on CRC only use the union fusion. In our MPFN model, we design another two fusion methods to extend the perspective of fusion. We evaluate the MPFN model on MRC task and achieve the state-of-the-art result.

\begin{figure}[!t]
\renewcommand{\arraystretch}{0.8}
\centering
\includegraphics[width=\textwidth, height=6cm]{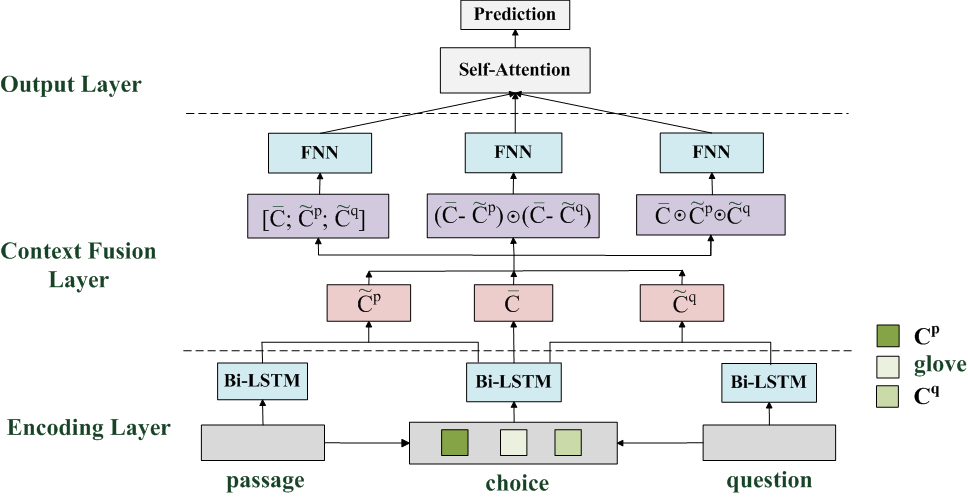}
\caption{\label{fig:model} Architecture of our MPFN  Model.}
\end{figure}

\section{Model}
The overview of our Multi-Perspective Fusion Network (MPFN) is shown in Fig.~\ref{fig:model}. 
Given a narrative passage about a series of daily activities and several corresponding questions, a system requires to select a correct choice from two options for each question. In this paper, we denote $\bf{p=\{p_1,p_2,...,p_{|p|}\}}$ as the passage, $\bf{q=\{q_1,q_2,...,q_{|q|}\}}$ as a question, $\bf{c=\{c_1,c_2,...,c_{|c|}\}}$ as one of the candidate choice, and a true label $y^{*} \in \{0,1\}$. Our model aims to compute a probability for each choice and take the one with higher probability as the prediction label. Our model consists of three layers: an encoding layer, a context fusion layer, and an output layer. The details of each layer are described in the following subsections. 

\subsection{Encoding Layer}
\label{section:encoding-layer}
This layer aims to encode the passage embedding $p$, the question embedding $q$, and the choice embedding $c$ into context embeddings. 
Specially, we use a one-layer BiLSTM as the context encoder.    
\begin{align}
&\bar{c}_i = \text{BiLSTM}(c, i) , & i \in [1,2,  \cdots,|c|]  \label{formula:c_i}\\
&\bar{p}_j = \text{BiLSTM}(p, j) , & j \in [1,2,  \cdots,|p|]  \label{formula:p_j} \\
&\bar{q}_k = \text{BiLSTM}(q, k) , & k \in [1,2,  \cdots,|q|] \label{formula:q_k}
\end{align}

The embeddings of $p$, $q$ and $c$  are semantically rich word representations consisting of several kinds of embeddings. Specifically, the embeddings of passage and question are the concatenation of the Golve word embedding, POS embedding, NER embedding, Relation embedding and Term Frequency feature. And the embeddings of choice comprise the Golve word embedding, the choice-aware passage embedding, \deleted{$c^p$}and choice-aware question embedding \deleted{$c^q$}. The details about each embedding are follows: 

\textbf{Glove word embedding} We use the 300-dimensional Glove word embeddings
trained from 840B Web crawl data \cite{Pennington14glove:global}. The out-of-vocabulary words are initialized randomly. The embedding matrix are fixed during training.

\textbf{POS\&NER embedding} We leverage the Part-of-Speech (POS) embeddings and Named-Entity Recognition(NER) embeddings. The two embeddings \deleted{$c_i^{pos} \text{and} c_i^{ner}$} are randomly initialized \deleted{to 12d and 8d respectively,} and updated during training. 

\textbf{Relation embedding} Relations are extracted form ConceptNet. 
For each word in the choice, if it satisfies any relation with another word in the passage or the question, the corresponding relation will be taken out. If the relations between two words are multiple, we just randomly choose one. The relation embeddings \deleted{$c_i^{rel}$}are generated in the similar way of POS embeddings.
\deleted{randomly initialized and updated during training as well.}

\textbf{Term Frequency} Following \cite{Wang:2018:TriAN}, we introduce the term frequency feature to enrich the embedding of each word. The calculation is based on English Wikipedia. 

\textbf{Choice-aware passage embedding} The information in the passage that is relevant to the choice can help encode the choice \cite{WeissenbornWS:2017:FastQA}. 
To acquire the choice-aware passage embedding $c_i^p$, we utilize dot product between non-linear mappings of word embeddings to compute the attention scores for the passage \cite{LeeKP:2016:RaSoR}.  
\begin{align}
& c_i^p = Attn(c_i,\{p_j\}_1^{|p|}) = \sum_{j=1}^{|p|} {\alpha}_{ij} p_j \\
& {\alpha}_{ij}  \propto exp(S(c_i, p_j)), \quad S(c_i, p_j) =   {ReLU(W{c_i})}^{T} ReLU(W {p_j})
\end{align}

\textbf{Choice-aware question embedding} The choice relevant question information is also important for the choice. Therefore, we adopt the similar attention way as above to get the choice-aware question embedding $c_i^q=Attn(c_i, \{q_k\}_{1}^{|q|})$. 

The embeddings delivered to the BiLSTM are the concatenation the above components, where $p_j = [p_j^{glove}, p_j^{pos},p_j^{ner},p_j^{rel}, p_j^{tf} ]$, $c_i = [c_i^{glove}, c_i^{p},c_i^{q}]$, and $q_k = [q_k^{glove}, q_k^{pos}, q_k^{ner}, q_k^{rel},q_k^{tf} ]$.

\subsection{Context Fusion Layer}
This is the core layer of our MPFN model. \deleted{To take the union, different and similar information of the choice, passage, and question into consideration, three fusion functions are defined in this layer. }
In this layer, we define three fusion functions, which consider the union information, the different information, and the similar information of the choice, passage, and question.

Since we have obtained the choice context $\bar{c}_i$, the passage context $\bar{p}_j$, and the question context $\bar{q}_k$ in the encoding layer, we can calculate the choice-aware passage contexts $\tilde{c}^p_i$ and choice-aware question contexts $\tilde{c}^q_i$. Then we deliver them together with the choice contexts $\bar{c}_i$ to the three fusion functions.

\deleted{In this layer, we define three fusion functions to fuse the $\bar{c}_i$, $\tilde{c}^p_j$, and $\bar{c}^q_k$ simultaneously and multi-perspectively. The three fusion functions take the union information, the different information, and the similar information of the choice, passage, and question into consideration. 
To better integrate this information, we feed the three fusion outputs to  FNN for aggregation.}

\textbf{Choice-aware passage context} In this part, we calculate the choice-aware passage representations $\tilde{c}_i^p= \sum_{j}{\beta}_{ij} \bar{p}_j$. 
For model simplification, here we use dot product between choice contexts and passage contexts to compute the attention scores ${\beta}_{ij}$:
\begin{align}
&{\beta}_{ij}= \frac{exp({\bar{c}_i^T \bar{p}_j)}} {\sum\limits{_{j^\prime=1}^{|p|}exp(\bar{c}_i^T \bar{p}_{j^\prime})}} 
\end{align}

\textbf{Choice-aware question context} In a similar way as above, we get the choice-aware question context $\tilde{c}_i^q= \sum_{j}{\beta}_{ik} \bar{q}_k$. The ${\beta}_{ik}$ is the dot product of the choice context $\bar{c}_i$ and question context $\bar{q}_k$. 

\textbf{Multi-perspective Fusion} This is the key module in our MPFN model.
The goal of this part is to produce multi-perspective fusion representation for the choice $\bar{c}_i$, the choice-aware passage $\tilde{c}^p_i$, and the choice-aware question $\tilde{c}^q_i$.
In this paper, we define fusion in three perspectives: \emph{union}, \emph{difference}, and \emph{similarity}. 
Accordingly, we define three fusion functions to describe the three perspectives. The outputs and calculation  of the three functions are as follows:
\deleted{: concatenation $;$, element-wise dot product and element-wise subtraction. }
\deleted{$f^u$, $f^d$, and $f^s$} 
\deleted{All of the three fusion functions take the choice context, the choice-aware passage, and the choice-aware question as input.} 
\begin{align}
&u_i  =  [\bar{c}_i \, ; \tilde{c}_i^p \,; \tilde{c}^q_i] \label{formula:f^c},\\
&d_i  =  ( \bar{c}_i - \tilde{c}_i^p)\odot(\bar{c_i} - \tilde{c}_i^q) \label{formula:f^s},\\
&s_i  = \bar{c}_i \odot \tilde{c}_i^p \odot \tilde{c}_i^q \label{formula:f^m},
\end{align}
where $; \,$, $-$, and $\odot$ represent concatenation, element-wise subtraction, and element-wise multiplication respectively. And $u_i$, $d_i$, and $s_i$ are the representations from the union, difference and similarity perspective respectively.

The union perspective is commonly used in a large bulk of tasks \cite{Parikh:2016:DecomAtt,Huang:2017:FusionNet,xiong:2018:dcn-plus}.
It can see the whole picture of the passage, the question, and the choice by concatenating the $\tilde{c}^p_i$ and $\tilde{c}^q_i$ together with $c_i$ . While the difference perspective captures the different parts between choice and passage, and the difference parts between choice and question by $\bar{c_i} - \tilde{c}_i^p$ and $\bar{c_i} - \tilde{c}_i^q$ respectively.
The $\odot$ in difference perspective can detect the two different parts at the same time and emphasize them.  
In addition, the similarity perspective is capable of discovering the similar parts among the passage, the question, and the choice. 

To map the three fusion representations to lower and same dimension, we apply three different FNNs with the ReLU activation to $u_i$, $d_i$, and $s_i$. The final output $g_i$ is the concatenation of the results of the three FNNs, which represents a global perspective representation.
\begin{align} 
g_i=[f^u(u_i),f^d(d_i),f^s(s_i)] \label{formula:v_i}
\end{align}
\vspace{-27pt}
\subsection{Output Layer}
\vspace{-8pt}
The output layer includes a self-attention layer and a prediction layer. Following \cite{yang-EtAl:2016:N16-13}, we summarize the global perspective representation $\{g_i\}_1^{|c|}$ to a fixed length vector $r$. We compute the $r= \sum_{i=1}^{|c|} b_i g_i$, where $b_j$ is the self-weighted attention score : 
\begin{align}
&b_i =  \frac{exp(W{g}_i)}{\sum\limits{_{i^\prime=1}^{|c|}exp(W {g}_{i^\prime})}} 
\end{align}

In the prediction layer, we utilize the output of self-attention $r$ to make the final prediction. 

\deleted{The final output y is obtained by transforming the $\mathbf{v}$ to a scalar and then apply a sigmoid activation to map it to a probability. }
 
\section{Experiments}
\setcounter{footnote}{0}

\subsection{Experimental Settings}
%
%
\textbf{Data} We conduct experiments on the MCScript \cite{Ostermann:LREC18:MCScript}, which is used as the official dataset of SemEval2018 Task11. This dataset constructs a collection of text passages about daily life activities and a series of questions referring to each passage, and each question is equipped with two answer choices. 
The MCScript comprises 9731, 1411, and 2797 questions in training, development, and test set respectively. 
For data preprocessing, we use spaCy \footnote{https://github.com/explosion/spaCy} for sentence tokenization, Part-of-Speech tagging, and Name Entity Recognization. The relations between two words are generated by ConceptNet. 
\deleted{The MCScript is a recently released dataset, which collects 2,119 narrative texts about daily events along with 13,939 questions. In this dataset, 27.4\% questions require commonsense inference. }
\vspace{2pt}

\noindent\textbf{Parameters} We use the standard cross-entropy function as the loss function. 
We choose Adam \cite{Kingma2014AdamAM} with initial momentums for parameter optimization. 
As for hyper-parameters, we set the batch size as 32, the learning rate as 0.001, the dimension of BiLSTM and the hidden layer of FNN as 123. The embedding size of Glove, NER, POS, Relation are 300, 8, 12, 10 respectively.  The dropout rate of the word embedding and BiLSTM output are 0.386 and 0.40 respectively. 
\subsection{Experimental Results}
\begin{table}[t] 

\centering
\begin{tabular}{p{0.5\columnwidth}cc}
\toprule
Model & Test (\%acc) \\
\midrule  
SLQA   & 79.94\\
Rusalka & 80.48 \\
HMA Model (single)~\cite{Chen:2018:HMA} & 80.94\\
TriAN (single)~\cite{Wang:2018:TriAN}  & 81.94 \\ 
\textbf{MPFN} (single)& \bf{83.52}\\
\midrule
(jiangnan) (ensemble) \cite{Xia:2018:Jiangnan} & 80.91\\
MITRE (ensemble) \cite{Merkhofer:2018:MITRE}  & 82.27\\
TriAN (ensemble)~\cite{Wang:2018:TriAN}& 83.95\\
HMA Model (ensemble)~\cite{Chen:2018:HMA} &84.13\\
\textbf{MPFN} (ensemble)& \bf{84.84}\\
\bottomrule  
\end{tabular}\\[1em]
\caption{\label{table:main-result} Experimental Results of Models}
\end{table}


Table\ref{table:main-result} shows the results of our MPFN model along with the competitive models on the MCScript dataset. 
The TriAN achieves 81.94\% in terms of test accuracy, which is the best result of the single model.  
The best performing ensemble result is 84.13\%, provided by HMA, which is the voting results of 7 single systems.

Our single MPFN model achieves 83.52\% in terms of accuracy, outperforming all the previous models.
The model exceeds the HMA and TriAN by approximately 2.58\% and 1.58\% absolute respectively.
Our ensemble model surpasses the current state-of-the-art model with an accuracy of 84.84\%. 
We got the final ensemble result by voting on 4 single models. Every single model uses the same architecture but different parameters. 



\begin{table}[!t]
\noindent\parbox[t]{.5\linewidth}{%
\centering 
\begin{tabular}[t]{p{0.2\columnwidth}cc}
\toprule
Perspective & MPFN & MPFN+BiLSTM \\
\midrule  
U& 82.73 & 82.73\\
D & 82.27 & 81.77\\
S & 81.55 & 80.59\\
\midrule 
DU  & 82.84 & 82.16\\
SU & 82.48 & 82.87 \\
SD & 83.12 & 83.09 \\
\midrule 
SDU & \bf{83.52} & 82.70  \\
\bottomrule  
\end{tabular}\\[1em]
\caption{\label{table:7-perspective}Test Accuracy of Multi-Perspective}
}
\parbox[t]{.5\linewidth}{
\centering
\begin{tabular}[t]{p{0.24\columnwidth}cc}
\toprule
Model & Test (\%acc)\\
\midrule 
\textbf{MPFN}& \textbf{83.52}\\
\midrule
w/o POS & 82.70\\
w/o NER & 82.62\\
w/o Rel & 81.98\\
w/o TF & 81.91\\
\midrule  
w/o \text{$C^p$} & 81.62\\
w/o \text{$C^q$}  & 82.16\\
w/o \text{$C^p$\&$C^q$} & 81.66\\
\bottomrule
\end{tabular}\\ [1em]
\caption{\label{table:ablation-study}Encoding Inputs Ablation Study.}
}
\vspace{-9pt}
\end{table}



\subsection{Discussion of Multi-Perspective}
To study the effectiveness of each perspective, we conduct several experiments on the three single perspectives and their combination perspective. Table~\ref{table:7-perspective} presents their comparison results. The first group of models are based on the three single perspectives, and we can observe that the union perspective performs best compared with the difference and similarity perspective.  Moreover, the union perspective achieves 82.73\% in accuracy, exceeding the TriAN by 0.79\% absolute. We can also see that the similarity perspective is inferior to the other two perspectives.    

The second group of models \deleted{in the Table~\ref{table:7-perspective}}are formed from two perspectives. Compared with the single union perspective, combining the difference perspective with the union perspective can improve 0.11\%. Composing union and similarity fusion together doesn't help the training. To our surprise, the combination of similarity perspective and difference perspective obtains 83.09\% accuracy score.

The last model is our MPFN model, which performing best.
The final result indicates that composing the union perspective, difference perspective, and similarity perspective together to train is helpful.

Many advanced models employ a BiLSTM to further aggregate the fusion results. To investigate whether a BiLSTM can assist the model, we apply another BiLSTM to the three fusion representations in Formula~\ref{formula:v_i} respectively and then put them together. The results are shown in the second column in Table~\ref{table:7-perspective}, which indicate that the BiLSTM does not help improve the performance of the models. 

\begin{figure}[!tbp]
\centering
\includegraphics[width=\textwidth,height=6cm]{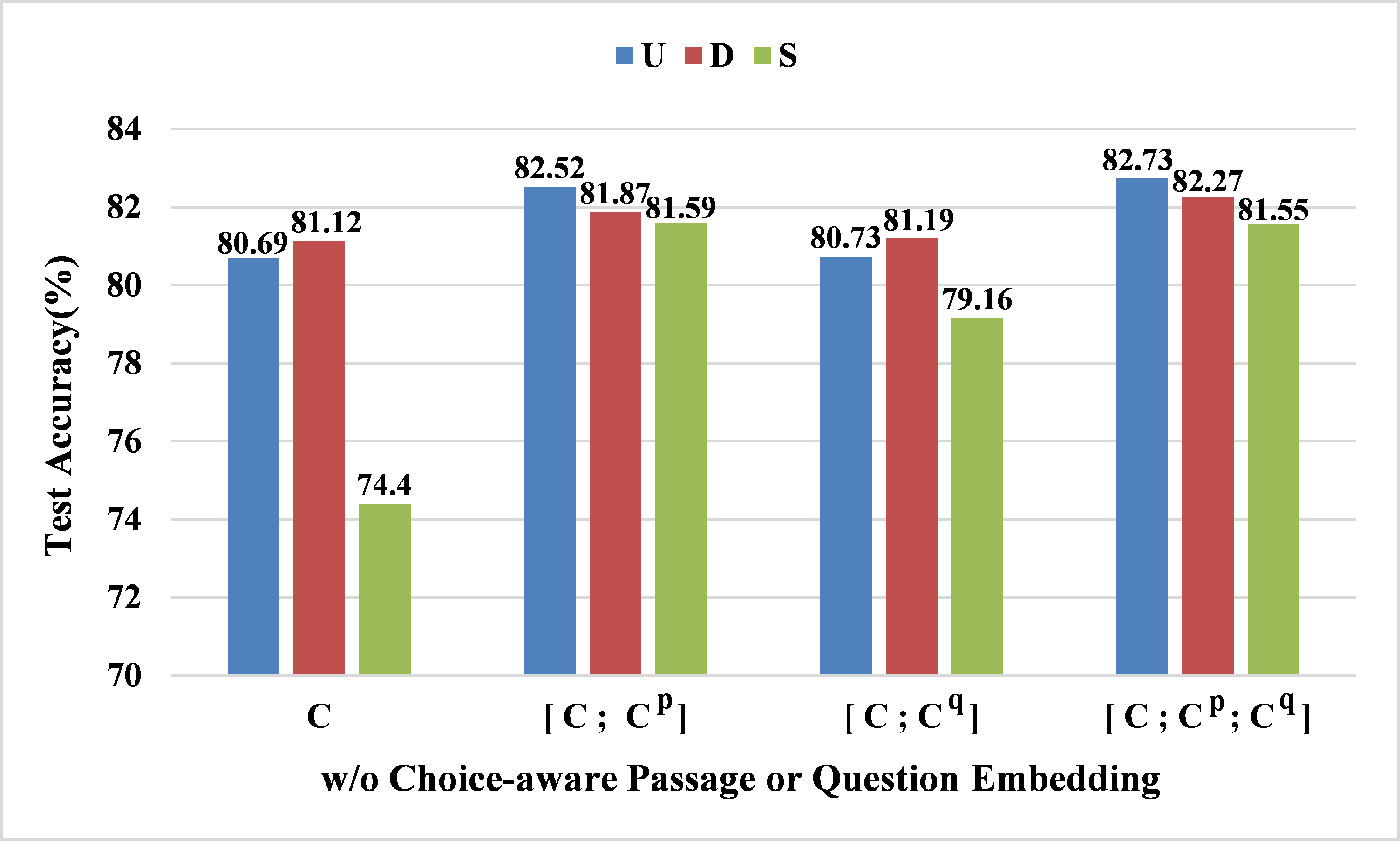}
\caption{\label{fig:word-level-interaction} Influence of Word-level Interaction.}
\end{figure}
\subsection{Encoding Inputs Ablation}

In the section, we conduct ablation study on the encoding inputs to examine the effectiveness each component. The experiment results are listed in Table~\ref{table:ablation-study}. 
\deleted{
In Section~\ref{section:encoding-layer}, we describe that our encoding inputs comprise six components: POS embedding, NER embedding, Relation embedding, Term Frequency, choice-aware passage embedding \text{$C^p$} and choice-aware question embedding \text{$C^q$}. 
}

From the best model, if we remove the POS embedding and NER embedding, the accuracy drops by 0.82\% and 0.9\%. 
Without Relation embedding, the accuracy drops to 81.98\%, revealing that the external relations are helpful to the context fusions. Without Term Frequency, the accuracy drops by approximately 1.61\%.  
This behavior suggests that the Term Frequency feature has a powerful capability to guide the model.

After removing the \text{$C^p$}, we find the performance
degrades to 81.62\%. This demonstrates that information in the passage is significantly important to final performance. If we remove \text{$C^q$} from the MPFN, the accuracy drops to 82.16\%. If we remove the word level fusion completely, we will obtain an 81.66\% accuracy score. 
These results demonstrate that each component is indispensable  and the bottom embeddings are the basic foundations of the top layer fusions.

\subsection{Influence of Word-level Interaction}
In this section, we explore the influence of word-level interaction to each perspective. 
Fig~\ref{fig:word-level-interaction} reports the overall results of how each perspective can be affected by the lower level interaction. 
The \text{$C^p$} and the \text{$C^q$} represent the choice-aware passage embedding and the choice-aware question embedding respectively. 
We can observe that the results of \text{$[C;C^p]$}, \text{$[C;C^q]$}, and \text{$[C;C^p;C^q]$} are all higher than the result of \text{$C$} alone, indicating the effectiveness of word embedding interaction. 

Both the union fusion and difference fusion can achieve more than 80\% accuracy, while the similarity fusion is very unstable.
We also observe that the difference fusion is comparable with the union fusion, which even works better than the union fusion when the information of \text{$C^p$} is not introduced into the input of encoding. The similarity fusion performs poorly in \text{$C$} and \text{$[C;C^q]$}, while yielding a huge increase in the remaining two groups of experiments, which is an interesting phenomenon. We infer that the similarity fusion needs to be activated by the union fusion.

In summary, we can conclude that integrate the information of \text{$C^p$} into \text{$C$} can greatly improve the performance of the model. Combining \text{$C^q$} together with \text{$C^p$} can further increase the accuracy. 
\deleted{
The information in the passage is richer than the question
The overall conclusion 
}

\subsection{Visualization}
\begin{figure}[htbp]
\centering
\begin{minipage}[t]{0.48\textwidth}
\centering
\includegraphics[scale=0.38]{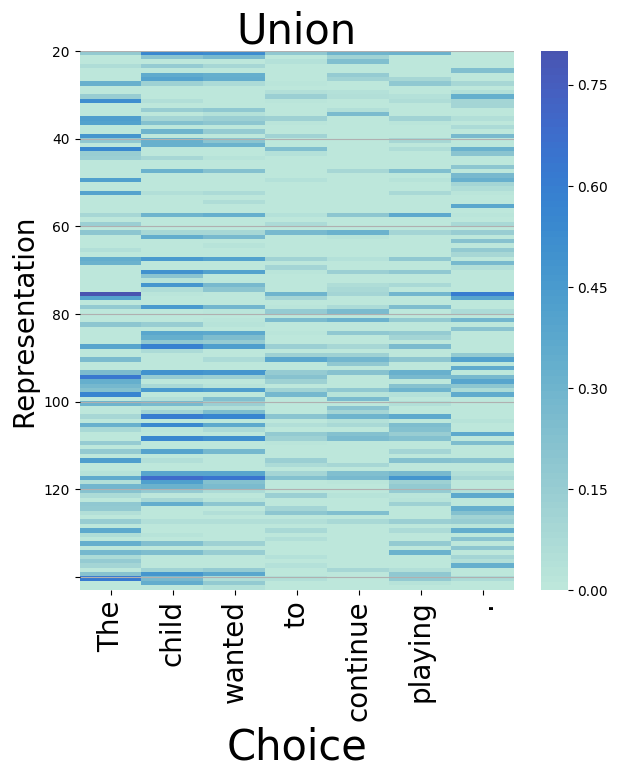}
\end{minipage}
\begin{minipage}[t]{0.48\textwidth}
\centering
\includegraphics[scale=0.38]{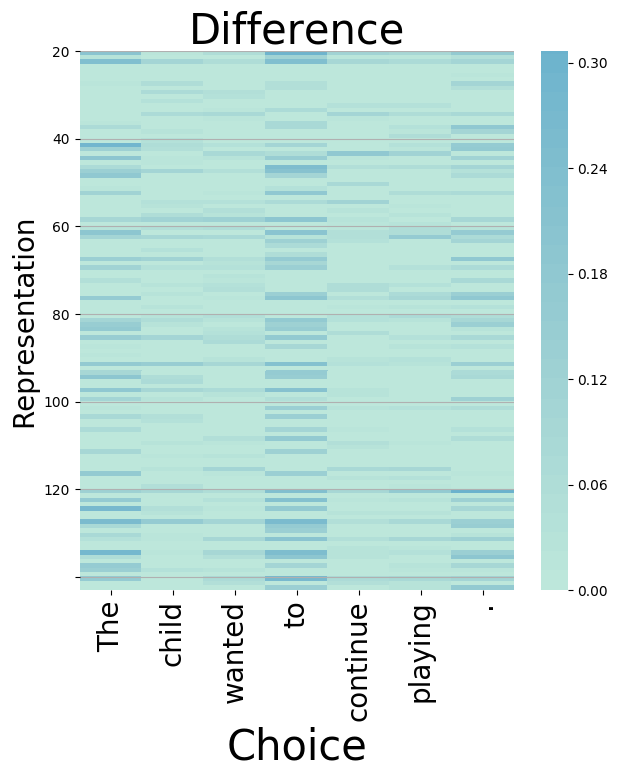}
\end{minipage}
\caption{Visualization of Fusions}
\label{fig:fusion-visualzation}
\end{figure}

In this section, we visualize the union and difference fusion representations and show them in Fig~\ref{fig:fusion-visualzation}. And, we try to analyze their characteristics and compare them to discover some connections. The values of similarity fusion are too small to observe useful information intuitively, so we do not show it here. 
We use the example presented in Table~\ref{table:example_MCS} for visualization, where the question is \emph{Why didn't the child go to bed by themselves?} and the corresponding True choice is \emph{The child wanted to continue playing.}

The left region in Fig~\ref{fig:fusion-visualzation} is the union fusion. The most intuitive observation is that it captures comprehensive information. The values of \emph{child}, \emph{wanted}, \emph{playing} are obvious higher than other words. This is consistent with our prior cognition, because the concatenation operation adopted in union fusion does not lose any content. 
While the difference union shows in the right region in Fig~\ref{fig:fusion-visualzation} focuses on some specific words. By further comparison, we find that the difference fusion can pay attention to the content ignored by the union fusion. What's more, the content acquired by the union would not be focused by the difference again. 
In other words, the union fusion and difference fusion indeed can emphasize information from the different perspective.
\deleted{Due to space limitation and} 


\deleted{\subsection{Error Analysis}
  
}

\section{Conclusion}
In this paper, we propose the Multi-Perspective Fusion Network (MPFN) for the Commonsense Reading Comprehension (CMC) task. 
We propose a more general framework for CRC by designing the difference and similarity fusion to assist the union fusion. 
Our MPFN model achieves an accuracy of 83.52\% on MCScript, outperforming the previous models. The experimental results show that union fusion based on the choice-aware passage, the choice-aware question, and the choice can surpass the TriAN and HMA model. The difference fusion performs stably, which is comparable with the union fusion.   
We find that the word-level union fusion can significantly influence the context-level fusion. The choice-aware passage word embedding can activate the similarity fusion. We find that combining the similar parts and the difference parts together can obtain the best performance among the two-perspective models. 
By taking the three types of fusion methods into consideration, our MPFN model achieves a state-of-the-art result.  

\section*{Acknowledgements}
This work is funded by Beijing Advanced Innovation for Language Resources of BLCU, the Fundamental Research Funds for the Central Universities in BLCU (17PT05), the Natural Science Foundation of China (61300081), and the Graduate Innovation Fund of BLCU (No.18YCX010).
%
%
%
\bibliographystyle{splncs04}

\bibliography{ccl2018}

\begin{thebibliography}{10}
\providecommand{\url}[1]{\texttt{#1}}
\providecommand{\urlprefix}{URL }
\providecommand{\doi}[1]{https://doi.org/#1}

\bibitem{Chen-zhu:2017:Long3}
Chen, Q., Zhu, X., Ling, Z.H., Wei, S., Jiang, H., Inkpen, D.: Enhanced lstm
  for natural language inference. In: Proceedings of the 55th Annual Meeting of
  the Association for Computational Linguistics (Volume 1: Long Papers). pp.
  1657--1668. Association for Computational Linguistics, Vancouver, Canada
  (July 2017), \url{http://aclweb.org/anthology/P17-1152}

\bibitem{Chen:2018:HMA}
Chen, Z., Cui, Y., Ma, W., Wang, S., Liu, T., Hu, G.: Hfl-rc system at
  semeval-2018 task 11: Hybrid multi-aspects model for commonsense reading
  comprehension. arXiv preprint arXiv:1803.05655  (2018)

\bibitem{Merkhofer:2018:MITRE}
Elizabeth M.~Merkhofer, John~Henderson, D.B.L.S., Zarrella, G.: Mitre at
  semeval-2018 task 11: Commonsense reasoning without commonsense knowledge
  (2018)

\bibitem{Hochreiter:1997:LSTM}
Hochreiter, S., Schmidhuber, J.: Long short-term memory. Neural Comput.
  \textbf{9}(8),  1735--1780 (Nov 1997). \doi{10.1162/neco.1997.9.8.1735},
  \url{http://dx.doi.org/10.1162/neco.1997.9.8.1735}

\bibitem{Huang:2017:FusionNet}
Huang, H., Zhu, C., Shen, Y., Chen, W.: Fusionnet: Fusing via fully-aware
  attention with application to machine comprehension. CoRR
  \textbf{abs/1711.07341} (2017)

\bibitem{Joshi:2017:TriviaQA}
Joshi, M., Choi, E., Weld, D.S., Zettlemoyer, L.: Triviaqa: A large scale
  distantly supervised challenge dataset for reading comprehension. In:
  Proceedings of the 55th Annual Meeting of the Association for Computational
  Linguistics. Association for Computational Linguistics, Vancouver, Canada
  (July 2017)

\bibitem{Kingma2014AdamAM}
Kingma, D.P., Ba, J.: Adam: A method for stochastic optimization. CoRR
  \textbf{abs/1412.6980} (2014)

\bibitem{Lai:2017:RACE}
Lai, G., Xie, Q., Liu, H., Yang, Y., Hovy, E.H.: Race: Large-scale reading
  comprehension dataset from examinations. In: EMNLP (2017)

\bibitem{LeeKP:2016:RaSoR}
Lee, K., Kwiatkowski, T., Parikh, A.P., Das, D.: Learning recurrent span
  representations for extractive question answering. CoRR
  \textbf{abs/1611.01436} (2016), \url{http://arxiv.org/abs/1611.01436}

\bibitem{Mou-EtAl:2016:P16-2}
Mou, L., Men, R., Li, G., Xu, Y., Zhang, L., Yan, R., Jin, Z.: Natural language
  inference by tree-based convolution and heuristic matching. In: Proceedings
  of the 54th Annual Meeting of the Association for Computational Linguistics
  (Volume 2: Short Papers). pp. 130--136. Association for Computational
  Linguistics, Berlin, Germany (August 2016),
  \url{http://anthology.aclweb.org/P16-2022}

\bibitem{Ostermann:LREC18:MCScript}
Ostermann, S., Modi, A., Roth, M., Thater, S., Pinkal, M.: {MCScript: A Novel
  Dataset for Assessing Machine Comprehension Using Script Knowledge}. In:
  Proceedings of the Eleventh International Conference on Language Resources
  and Evaluation (LREC 2018). European Language Resources Association (ELRA),
  Miyazaki, Japan (May 7-12, 2018 2018)

\bibitem{Parikh:2016:DecomAtt}
Parikh, A., T{\"a}ckstr{\"o}m, O., Das, D., Uszkoreit, J.: A decomposable
  attention model for natural language inference. In: Proceedings of the 2016
  Conference on Empirical Methods in Natural Language Processing. pp.
  2249--2255. Association for Computational Linguistics (2016).
  \doi{10.18653/v1/D16-1244}, \url{http://www.aclweb.org/anthology/D16-1244}

\bibitem{Pennington14glove:global}
Pennington, J., Socher, R., Manning, C.D.: Glove: Global vectors for word
  representation. In: In EMNLP (2014)

\bibitem{Rajpurkar:2016:SQUAD}
Rajpurkar, P., Zhang, J., Lopyrev, K., Liang, P.: Squad: 100, 000+ questions
  for machine comprehension of text. CoRR  \textbf{abs/1606.05250} (2016)

\bibitem{Richardson:2013:MCTestAC}
Richardson, M., Burges, C.J.C., Renshaw, E.: Mctest: A challenge dataset for
  the open-domain machine comprehension of text. In: EMNLP (2013)

\bibitem{Seo:2016:BiDAF}
Seo, M.J., Kembhavi, A., Farhadi, A., Hajishirzi, H.: Bidirectional attention
  flow for machine comprehension. CoRR  \textbf{abs/1611.01603} (2016)

\bibitem{Trischler:2017:NewsQA}
Trischler, A., Wang, T., Yuan, X., Harris, J., Sordoni, A., Bachman, P.,
  Suleman, K.: Newsqa: A machine comprehension dataset. In: Rep4NLP@ACL (2017)

\bibitem{Wang:2018:TriAN}
Wang, L., Sun, M., Zhao, W., Shen, K., Liu, J.: Yuanfudao at semeval-2018 task
  11: Three-way attention and relational knowledge for commonsense machine
  comprehension. In: SemEval@NAACL-HLT. pp. 758--762. Association for
  Computational Linguistics (2018)

\bibitem{wang-jiang:2016:mlstm}
Wang, S., Jiang, J.: Learning natural language inference with lstm. In:
  Proceedings of the 2016 Conference of the North American Chapter of the
  Association for Computational Linguistics: Human Language Technologies. pp.
  1442--1451. Association for Computational Linguistics, San Diego, California
  (June 2016), \url{http://www.aclweb.org/anthology/N16-1170}

\bibitem{Wang:2017:r-net}
Wang, W., Yang, N., Wei, F., Chang, B., Zhou, M.: Gated self-matching networks
  for reading comprehension and question answering. In: Proceedings of the 55th
  Annual Meeting of the Association for Computational Linguistics (Volume 1:
  Long Papers). pp. 189--198. Association for Computational Linguistics (2017).
  \doi{10.18653/v1/P17-1018}, \url{http://www.aclweb.org/anthology/P17-1018}

\bibitem{WeissenbornWS:2017:FastQA}
Weissenborn, D., Wiese, G., Seiffe, L.: Fastqa: {A} simple and efficient neural
  architecture for question answering. CoRR  \textbf{abs/1703.04816} (2017),
  \url{http://arxiv.org/abs/1703.04816}

\bibitem{Weston:2015:BAbI}
Weston, J., Bordes, A., Chopra, S., Mikolov, T.: Towards ai-complete question
  answering: {A} set of prerequisite toy tasks. CoRR  \textbf{abs/1502.05698}
  (2015)

\bibitem{Xia:2018:Jiangnan}
Xia, J.: Jiangnan at semeval-2018 task 11: Deep neural network with attention
  method for machine comprehension task  (2018)

\bibitem{xiong:2018:dcn-plus}
Xiong, C., Zhong, V., Socher, R.: {DCN}+: Mixed objective and deep residual
  coattention for question answering. In: International Conference on Learning
  Representations (2018), \url{https://openreview.net/forum?id=H1meywxRW}

\bibitem{Xu:2017:DFN}
Xu, Y., Liu, J., Gao, J., Shen, Y., Liu, X.: Towards human-level machine
  reading comprehension: Reasoning and inference with multiple strategies. CoRR
   \textbf{abs/1711.04964} (2017)

\bibitem{yang-EtAl:2016:N16-13}
Yang, Z., Yang, D., Dyer, C., He, X., Smola, A., Hovy, E.: Hierarchical
  attention networks for document classification. In: NAACL. pp. 1480--1489.
  Association for Computational Linguistics, San Diego, California (June 2016),
  \url{http://www.aclweb.org/anthology/N16-1174}

\bibitem{Zhu:2018:RACE}
Zhu, H., Wei, F., Qin, B., Liu, T.: Hierarchical attention flow for
  multiple-choice reading comprehension. In: AAAI (2018)

\end{thebibliography}

%
\end{document}